\documentclass[letterpaper, 10 pt, journal, twoside]{IEEEtran}
\usepackage{times}
\usepackage[pdftex]{graphicx}

\usepackage{amsmath,amssymb,amsopn,amstext,amsfonts, amsthm}
\usepackage[space]{cite}
\usepackage{pdfsync}
\usepackage{balance}
\usepackage{color}
\usepackage{mathtools}
\usepackage[ruled,norelsize, linesnumbered]{algorithm2e}
\usepackage{bm}
\usepackage{float}
\usepackage[outdir=./]{epstopdf}
\usepackage{url}
\usepackage{multirow}
\usepackage{makecell}
\usepackage{adjustbox}
\usepackage{tabularx, booktabs}

\usepackage{enumitem}
\usepackage[font=small,skip=2pt]{caption}
\usepackage[linkcolor=black,citecolor=black,urlcolor=black,colorlinks=true]{hyperref}
\usepackage{subcaption}
\usepackage{bookmark}
\usepackage{tikz}
\usepackage{textcomp}

\DeclareGraphicsExtensions{.pdf}

\theoremstyle{remark}

\newcommand{\func}{\mathtt}

\newcommand{\set}{\mathcal}

\SetKw{Continue}{continue}
\makeatletter
\newcommand{\removelatexerror}{\let\@latex@error\@gobble}
\makeatother

\newcommand\copyrighttext{%
  \footnotesize \textcopyright 2019 IEEE. Accepted by IEEE RA-L. Personal use of this material is permitted. Permission from IEEE must be obtained for all other uses. }
\newcommand\copyrightnotice{%
\begin{tikzpicture}[remember picture,overlay]
\node[anchor=south,yshift=10pt] at (current page.south) {\fbox{\parbox{\dimexpr\textwidth-\fboxsep-\fboxrule\relax}{\copyrighttext}}};
\end{tikzpicture}%
}

\begin{document}

\title{Safe Trajectory Generation for Complex Urban\\ Environments Using Spatio-temporal\\ Semantic Corridor}

\author{Wenchao Ding$^1$$^\dagger$, Lu Zhang$^1$$^\dagger$, Jing Chen$^2$, and Shaojie Shen$^1$
\thanks{Accepted final version. To Appear in IEEE Robotics and Automation Letters. \textcopyright 2019 IEEE. Personal use of this material is permitted. Permission from IEEE must be obtained for all other uses. This work was supported by Hong Kong PhD Fellowship Scheme, HKUST-DJI Joint Innovation Laboratory, and HKUST Institutional Fund. (\textit{Corresponding author: Wenchao Ding.})}
\thanks{$^\dagger$W. Ding and L. Zhang contributed equally to this work. $^1$W. Ding, L. Zhang, and S. Shen are with the Department of Electronic and Computer Engineering, Hong Kong University of Science and Technology, Hong Kong 852, China (e-mail: wdingae@ust.hk; lzhangbz@ust.hk; eeshaojie@ust.hk).}
\thanks{$^2$J. Chen is with  DJI Technology Co., Ltd., Shenzhen 510810, China (e-mail: jing.chen@dji.com).}
}

\markboth{Author's Version}
{DING \MakeLowercase{\textit{et al.}}: Safe Trajectory Generation for Complex Urban Environments Using Spatio-temporal Semantic Corridor}

\maketitle

\begin{abstract}
	Planning safe trajectories for autonomous vehicles in complex urban environments is challenging since there are numerous semantic elements (such as dynamic agents, traffic lights and speed limits) to consider. These semantic elements may have different mathematical descriptions such as obstacle, constraint and cost. It is non-trivial to tune the effects from different combinations of semantic elements for a stable and generalizable behavior. In this paper, we propose a novel unified spatio-temporal semantic corridor (SSC) structure, which provides a level of abstraction for different types of semantic elements. The SSC consists of a series of mutually connected collision-free cubes with dynamical constraints posed by the semantic elements in the spatio-temporal domain. The trajectory generation problem then boils down to a general quadratic programming (QP) formulation. Thanks to the unified SSC representation, our framework can generalize to any combination of semantic elements. Moreover, our formulation provides a theoretical \textit{guarantee} that the \textit{entire} trajectory is safe and constraint-satisfied, by using the convex hull and hodograph properties of piecewise B\'{e}zier curve parameterization. We also release the code of our method to accommodate benchmarking.
\end{abstract}


\copyrightnotice

\section{Introduction}\label{sec:introduction}
\IEEEPARstart{T}{rajectory} generation for autonomous vehicles (AVs) in complex urban environments is challenging since there are many semantic elements (e.g., dynamic agents, traffic lights, speed limits, stop signs, and lane geometry). Different types of semantic elements may have different mathematical descriptions such as obstacle, constraint and cost~\cite{ajanovic2018search}. It is non-trivial to tune the effects from different combinations of semantic elements so that the formulation can generalize well to all combinations of semantic elements~\cite{gu2015tunable}. Therefore, it is essential to describe diverse kinds of semantic elements in a unified way such that the type and combination of semantic elements do not affect the planning performance.

Apart from the representation issue of the semantic elements, another issue is how to~\textit{guarantee} the safety and feasibility of the generated trajectory. Most existing optimization-based~\cite{ziegler2014local,xu2012optimization} and lattice-based~\cite{ziegler2009spatiotemporal,mcnaughton2011motion,martin2010On} motion planners try to check or enforce constraints at a series of sampled points. However, these methods may fail to detect or resolve infeasible points between two sample points, and thus cannot provide safety guarantee for the entire trajectory.
\begin{figure}[t]
	\centering
	\begin{subfigure}[b]{0.44\textwidth}
		\includegraphics[width =\textwidth]{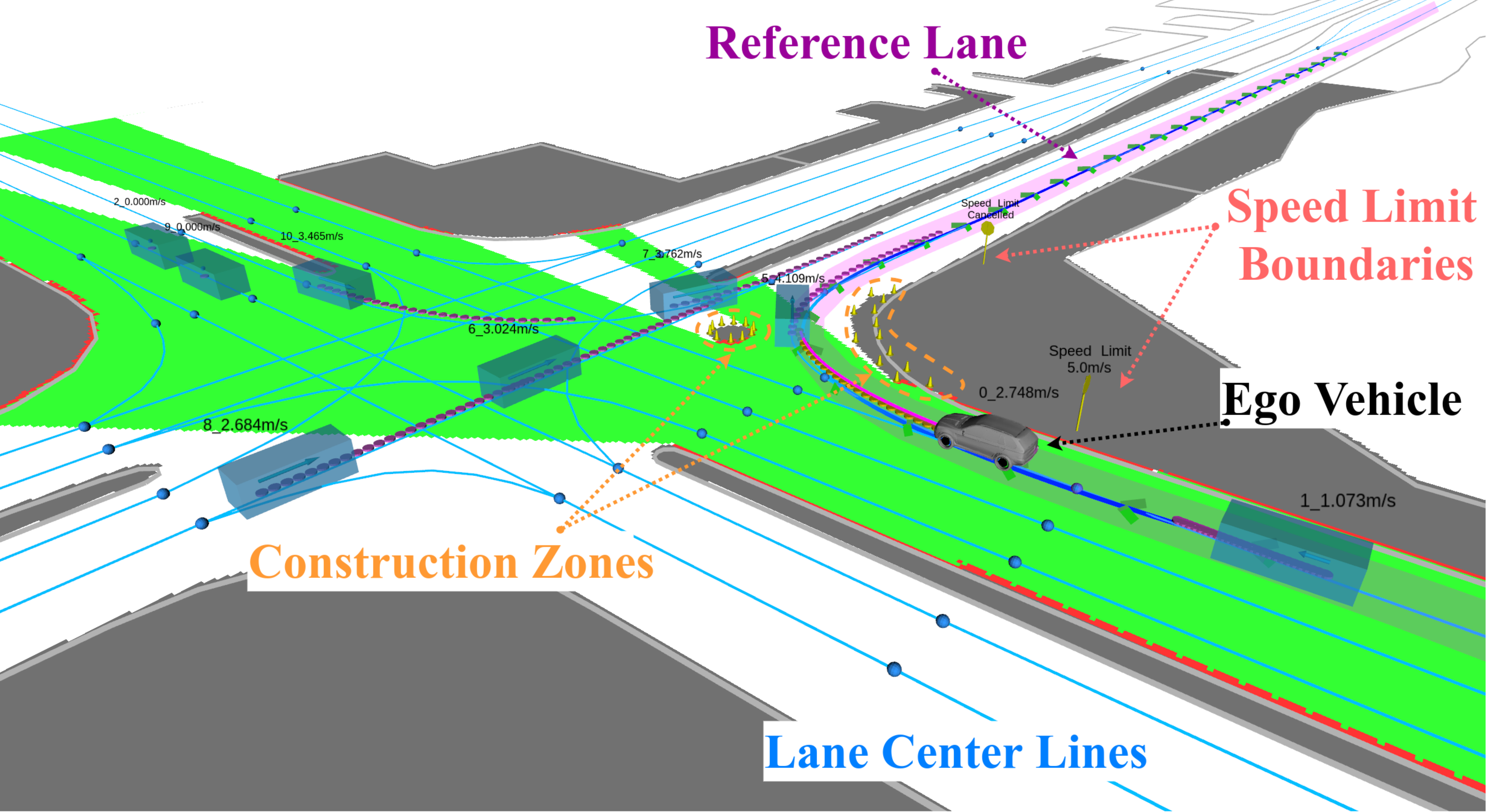}
		\caption{Complex urban driving environments}\label{fig:framework}
		\includegraphics[width =\textwidth]{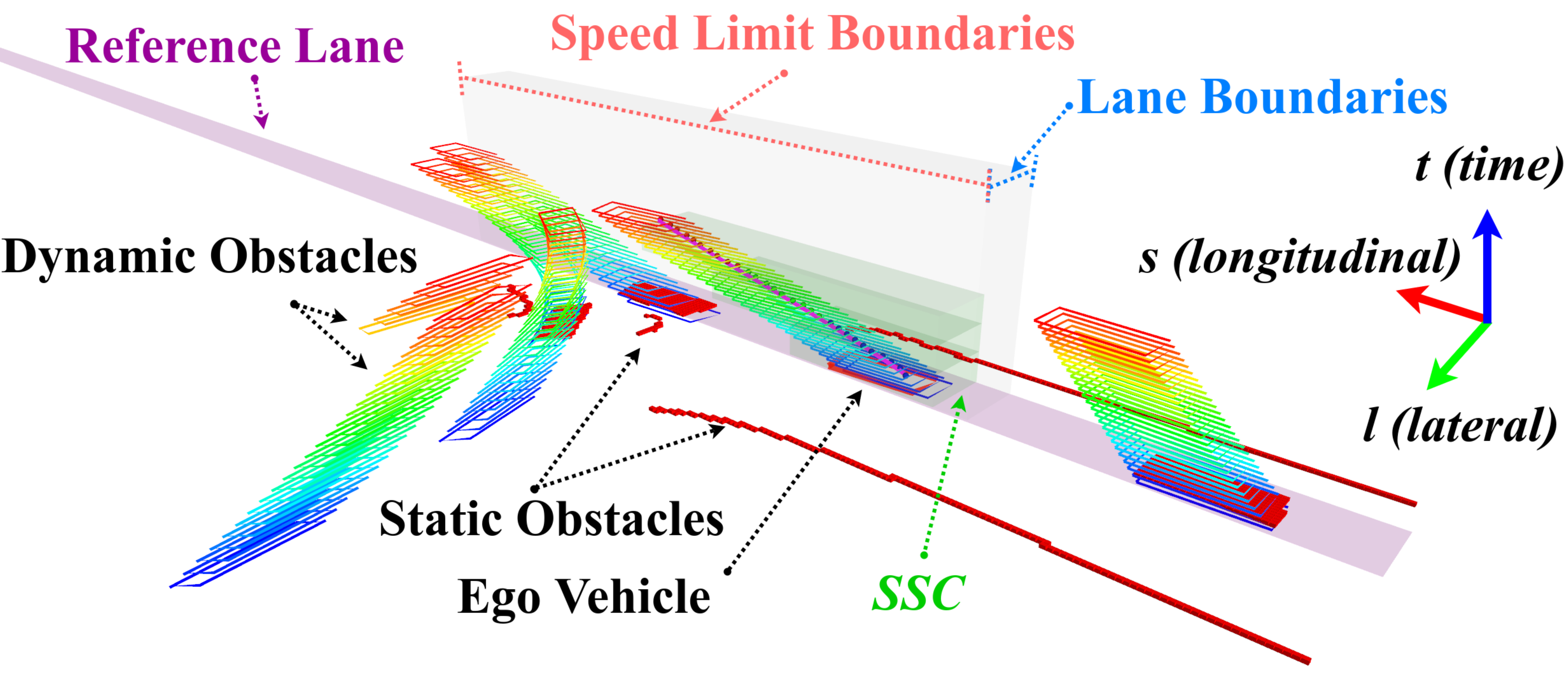}
		\caption{Spatio-temporal semantic corridor (SSC)}\label{fig:construction}
	\end{subfigure}%
	\caption{Illustration of our trajectory generation framework. Complex semantic elements of the environment are projected to the spatio-temporal domain w.r.t. the reference lane. The SSC encodes the requirements given by the semantic elements and a safe trajectory is generated accordingly. Note that the visualization of the static obstacles is clipped to show the details of other components. More examples can be found in the video \url{https://www.youtube.com/watch?v=LrGmKaM3Iqc}.}\label{fig:cover}
	\vspace{-0.5cm}
\end{figure}

To overcome the above challenges, we propose a unified trajectory generation framework with a theoretical safety and feasibility guarantee. The key to the framework is a novel spatio-temporal semantic corridor (SSC) structure. The SSC is motivated by the fact that most semantic elements can be either rendered as spatio-temporal obstacles or constraints within a certain range of the spatio-temporal domain. The key feature of the SSC is its abstraction for different types of semantic elements. Essentially, the SSC consists of a series of mutually connected collision-free cubes with dynamical constraints posed by the semantic elements. We propose an SSC generation process to generate and split the cubes so that the dynamical constraints can be correctly associated.

Given the unified SSC representation, the trajectory generation problem boils down to generating the optimal trajectory within the SSC while satisfying the dynamical constraints. In this paper, we contribute a quadratic programming (QP) formulation which guarantees the safety and feasibility of the generated trajectory by using piecewise B\'{e}zier curve parameterization. The proposed formulation is built on the top of the convex hull and hodograph properties of the B\'{e}zier curve. The contributions are summarized as follows:

\begin{itemize}
\item An SSC structure which provides a unified representation for diverse kinds of semantic elements in complex urban environments.
\item An optimization-based trajectory generation formulation which ensures safety and feasibility for the entire generated trajectory.
\item A complete and open-source trajectory generation framework and real-time implementation in a multi-agent urban simulation platform. Comprehensive experiments and comparisons are presented to validate the performance.
\end{itemize}

The related literature is reviewed in Sect.~\ref{sec:related_works}. An overview of our trajectory generation framework is provided in Sect.~\ref{sec:system_overview}. Our SSC generation method and trajectory generation method are detailed in Sect.~\ref{sec:ssc_generation} and Sect.~\ref{sec:traj_gen}, respectively. Experimental results and benchmark analysis are elaborated in Sect.~\ref{sec:experimental_results}. Finally, a conclusion is drawn in Sect.~\ref{sec:conclusion}.

\section{Related Works}\label{sec:related_works}
\subsection{Spatio-temporal motion planning for AVs}
There is extensive literature on spatio-temporal motion planning for autonomous vehicles. Ziegler~\textit{et al.}~\cite{ziegler2009spatiotemporal} sample a spatio-temporal state lattice on a space-time manifold~\cite{likhachev2009primitive} and use a search-based approach to obtain an executable trajectory. McNaughton~\textit{et al.}~\cite{mcnaughton2011motion} adopt a spatio-temporal state lattice, which can automatically conform to the lane geometry by numerical optimization. Due to an unacceptable blowup in the size of the search space (i.e., the \textit{curse of dimensionality}), GPU-accelerated dynamic programming is adopted in~\cite{mcnaughton2011motion}. However, the semantic elements in urban environments (such as speed limits, traffic lights, etc.) are not modeled in~\cite{ziegler2009spatiotemporal, mcnaughton2011motion, likhachev2009primitive}.

There are several approaches that attempt to model the semantic elements. Wolf~\textit{et al.}~\cite{wolf2008artificial} associate semantic elements with specially designed cost functions and aggregate the cost terms as a potential field. However, this approach suffers from local minimums. Moreover, it is non-trivial to correctly balance the effects of different cost terms for different configurations of the semantic elements\cite{gu2015tunable}. Hubmann~\textit{et al.}~\cite{hubmann2016generic} render traffic lights and dynamic agents as obstacles in the longitudinal and time domain and apply a search-based method to obtain a generic driving strategy (i.e., a speed plan). Ajanovic~\textit{et al.}~\cite{ajanovic2018search} extend the obstacle representation and render forbidden lane changes and solid lines as obstacles and speed limits as velocity constraints.

Built on top of the obstacle-like and constraint-like representations in~\cite{ajanovic2018search}, we further propose the SSC structure to generally represent different types of semantic elements. The key feature of the SSC is that it provides a level of abstraction which encodes all the information needed for later optimization. Adding a new semantic element or combining different semantic elements does not affect the cost formularization and constraint specification, which renders a unified and generalizable trajectory generation framework.

\subsection{Corridor generation for AVs}
The spatial corridor (i.e., convex free-space) is widely applied in trajectory generation. Zhu~\textit{et al.}~\cite{zhu2015convex} propose a convex elastic smoothing algorithm, which can generate a collision-free ``tube'' around the initial path and formulate the trajectory smoothing problem into a quadratically constrained quadratic programming (QCQP). Erlien~\textit{et al.}~\cite{erlien2013safe} consider not only spatial information but also vehicle dynamics to construct the convex tube. Both of these works, however, generate the corridor in a static environment and cannot deal with dynamic obstacles. Liu~\textit{et al.}~\cite{liu2017convex} find a convex feasible set around the reference trajectory and leverage the convex feasible set to accelerate the non-convex optimization. However, the computation complexity is still prohibitively high for real-time applications. Moreover, collision-avoidance is their major concern and semantic elements are not considered. We are motivated by these corridor generation methods and further extend the spatial corridor to the spatio-temporal domain to cope with dynamic obstacles. Additionally, the proposed SSC can take various kinds of semantic elements into account.

\begin{figure}[t]
	\centering
	\includegraphics[width=0.48\textwidth]{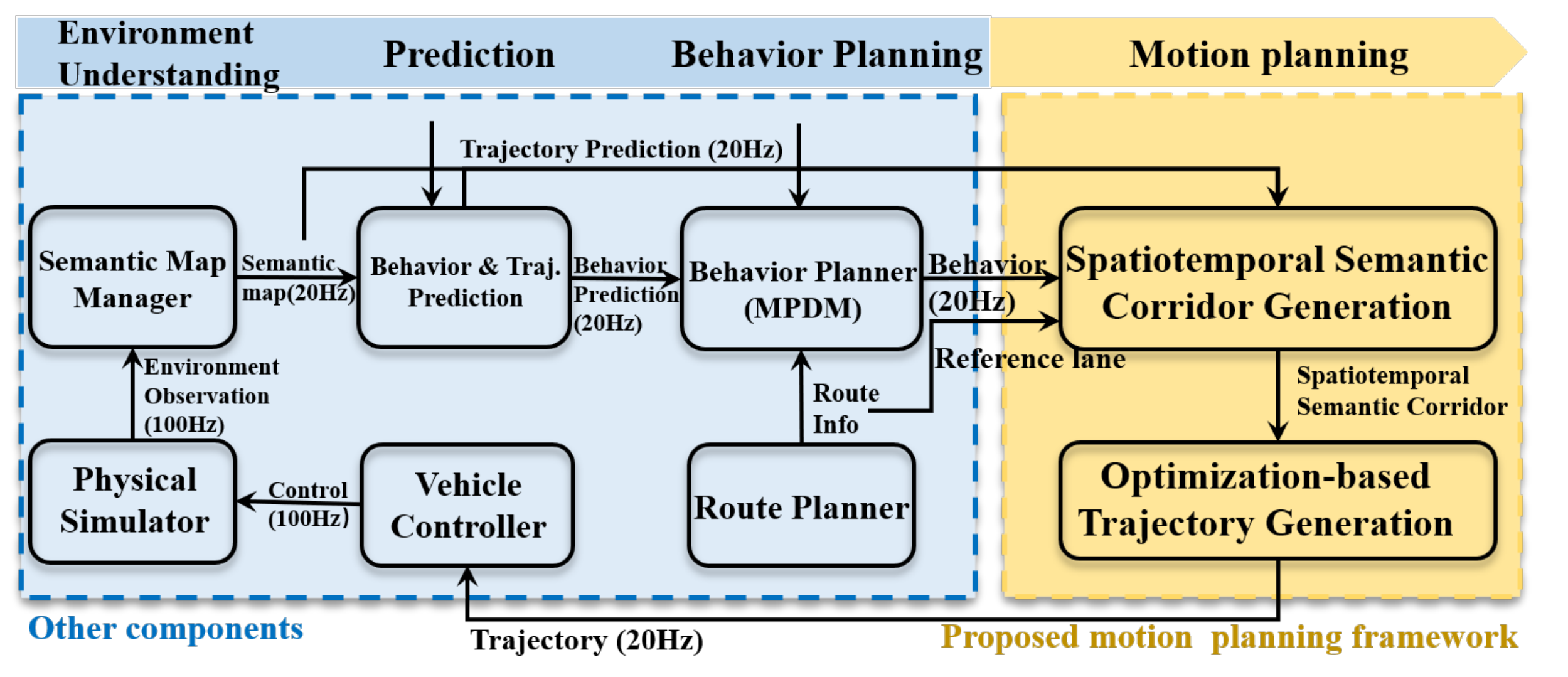}
	\caption{Illustration of the proposed trajectory generation framework and its relationship with other system components.}\label{fig:overview}
	\vspace{-0.5cm}
\end{figure}

\begin{figure*}[t]
  \centering
  \begin{subfigure}[b]{0.32\textwidth}
  	\centering
  	\includegraphics[width =0.95\textwidth]{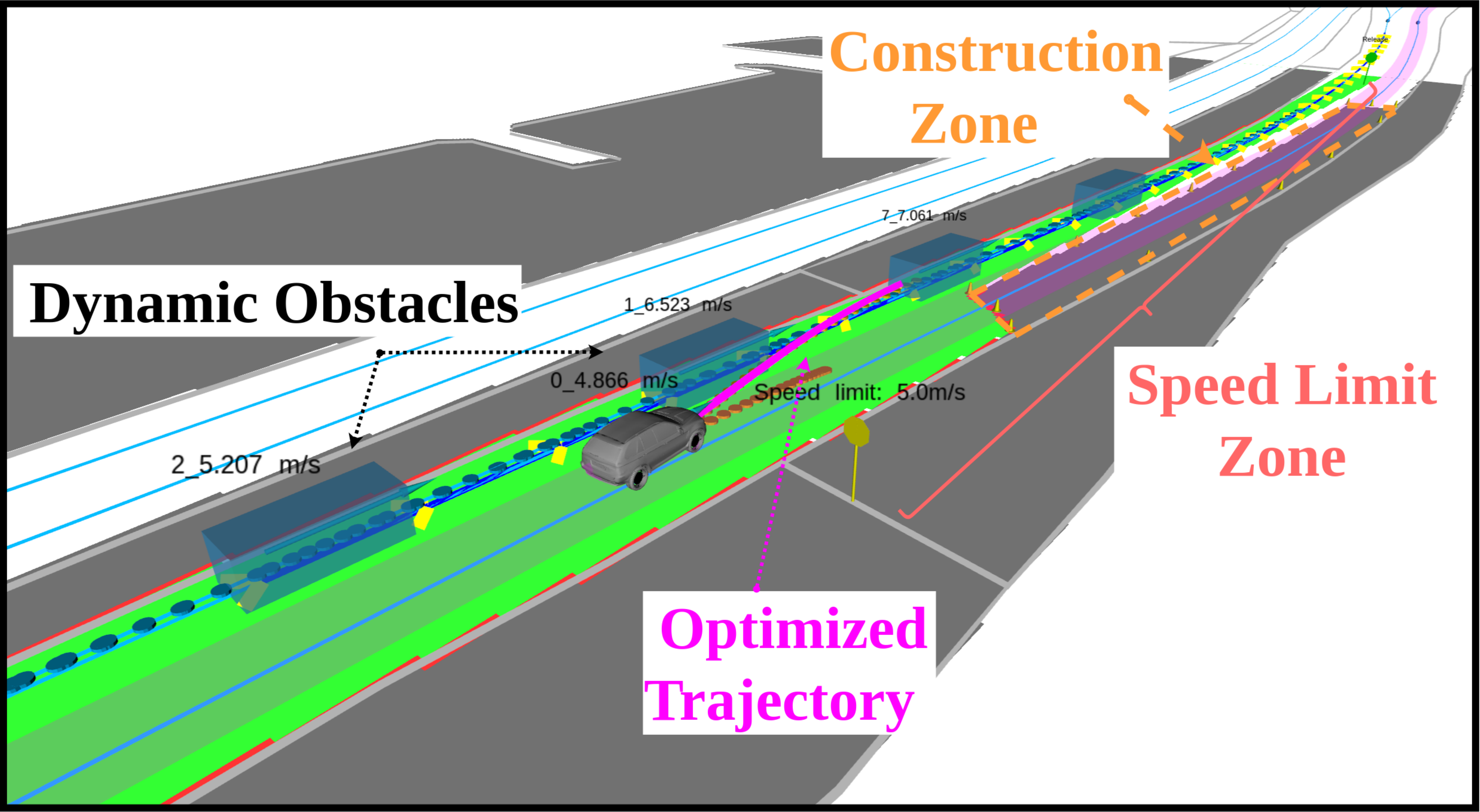}
  	\caption{Merging into congested traffic}\label{fig:workflow_merge}
  \end{subfigure}
  \begin{subfigure}[b]{0.32\textwidth}
  	\centering
    \includegraphics[width =0.95\textwidth]{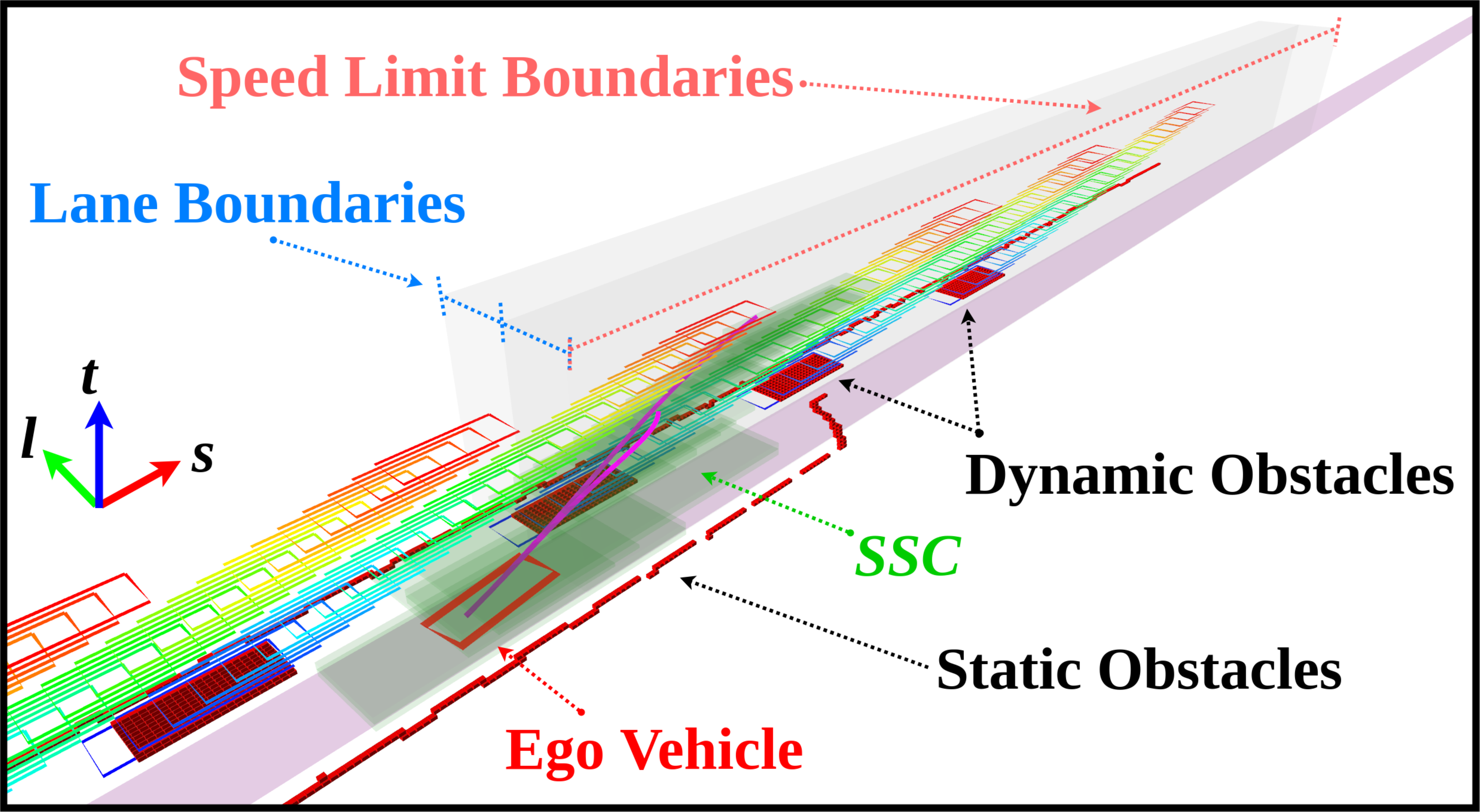}
    \caption{Spatio-temporal representation}\label{fig:workflow_slt}
  \end{subfigure}
  \begin{subfigure}[b]{0.32\textwidth}
  	\centering
    \includegraphics[width =0.95\textwidth]{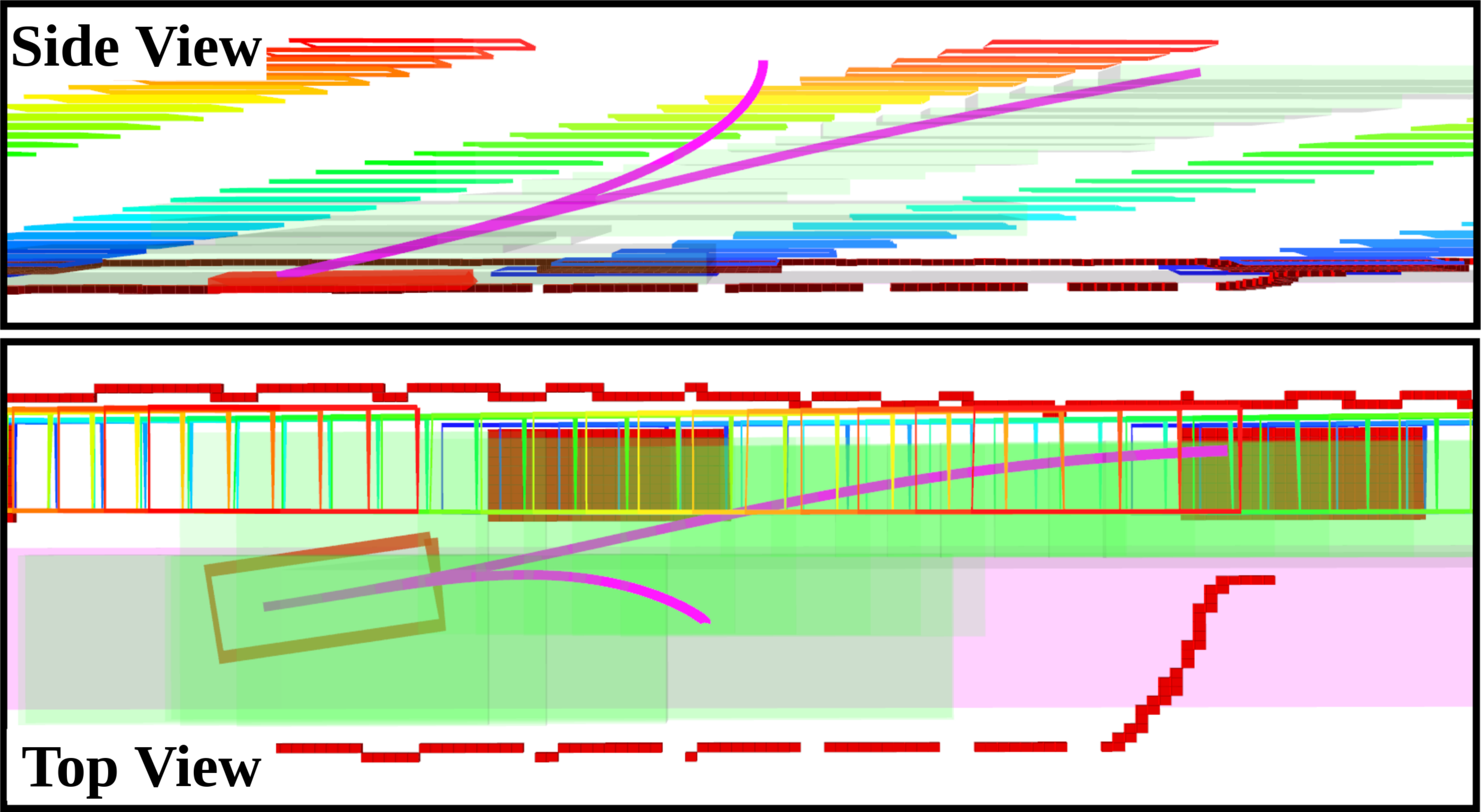}
    \caption{SSC and optimization}\label{fig:workflow_corridor}
  \end{subfigure}
	\caption{Illustration of merging into congested traffic under a speed limit. For the two potential behaviors, i.e., lane change and lane keeping, the optimal trajectories are generated inside the SSC for each behavior.}\label{fig:workflow}
	\vspace{-0.4cm}
\end{figure*}

\section{System Overview}\label{sec:system_overview}
The proposed trajectory generation framework (Fig.~\ref{fig:overview}) belongs to the motion planning layer of an autonomous vehicle, and it requires necessary inputs from the upper layers, e.g., the behavioral layer. Apart from the proposed trajectory generation, the other system components are also illustrated to clarify the input and output of our framework.

As depicted in Fig.~\ref{fig:overview}, there are four phases for a single planning cycle. The first phase is the environment understanding obtained by a semantic map manager which takes the responsibility to manage the semantic elements (e.g., occupancy grid map, dynamic agents, lanes, traffic rules, etc.) for local planning purposes. The second phase is the prediction, which not only provides high-level behavior anticipations (e.g., lane change, lane keeping, etc.) but also predicted trajectories for other dynamic agents. The third phase is the behavior planning, which is implemented using the multi-policy decision making (MPDM) method, as elaborated in Sect.~\ref{sec:preliminary_mpdm}. The fourth phase is our proposed motion planning, which takes discretized future simulated states from the behavior planner as seeds for corridor generation. Note that our trajectory generation framework can also work with other behavior planners, such as those in~\cite{hubmann2016generic, hubmann2017decision, chen2018continuous}, as long as the behavior planner provides a preliminary initial guess about the future states.

To construct the SSC, four ingredients are needed, namely, a semantic map which consists of the semantic elements, predicted trajectories for dynamical agents, forward simulated states, and a reference lane given by the route information. Note that the trajectory prediction module may be optional if the forward simulated states already include the states for other agents such as the case of MPDM. In such case, we can use the simulated states of other vehicles as the predicted trajectories, which facilitates passing interaction anticipations from behavior planning to motion planning layer. However, since this is not a common feature in behavior planning, we still use the predicted trajectories from the trajectory prediction module in the experiments for generality, which may lose the interaction information from behavior planning. To summarize, the source of the seeds and the modeling of interaction depend on the choice of behavior planner.

\section{Preliminaries on Multi-policy Decision Making}\label{sec:preliminary_mpdm}
In this paper, we adopt MPDM~\cite{galceran2015mpdmchangept} as the behavioral layer. Recall that our trajectory generation method can also work with other behavior planning methods~\cite{hubmann2016generic, hubmann2017decision, chen2018continuous}. Since behavior planning is out of the scope of this paper, only preliminary information about MPDM is provided here.

The MPDM model formulates the behavior planning problem as a general multi-agent partially observable Markov decision process (POMDP) to model the interaction and uncertainty in dynamic environments. Since solving the POMDP quickly becomes computationally intractable when the number of vehicles increases, MPDM relaxes the problem and assumes that both our vehicle and the other agents are executing a finite set of closed-loop discrete policies (e.g., lane change, lane keeping, etc.). Moreover, for each closed-loop policy, the future situation is anticipated via forward simulating all the vehicle states using a simplified simulation model, such as an idealized steering and speed controller. A comprehensive reward function is designed to assess the future situation and the best behavior is elected.

In this paper, we use the forward simulated states of the ego vehicle as the seeds in the corridor generation process. Although the initial seeds are collision-free, they can not be directly executed by the vehicle due to a coarse resolution ($0.15$ $s$ in the experiments) and a simplified simulation model (e.g., piecewise linear control in the experiments).

Since MPDM provides the forward simulated states for multiple behaviors (e.g., lane change left, lane change right, and lane keeping) at the same time, we fully utilize this feature and generate candidate trajectories for all the potential behaviors to enhance the robustness of the framework. For example, while executing a lane change trajectory, our trajectory framework always prepares the trajectory for switching back to the original lane, as shown in Fig.~\ref{fig:workflow}.

\begin{figure*}[t]
	\centering
	\includegraphics[trim=0.0cm 0.0cm 0.0cm 0.0cm, width=0.98\textwidth]{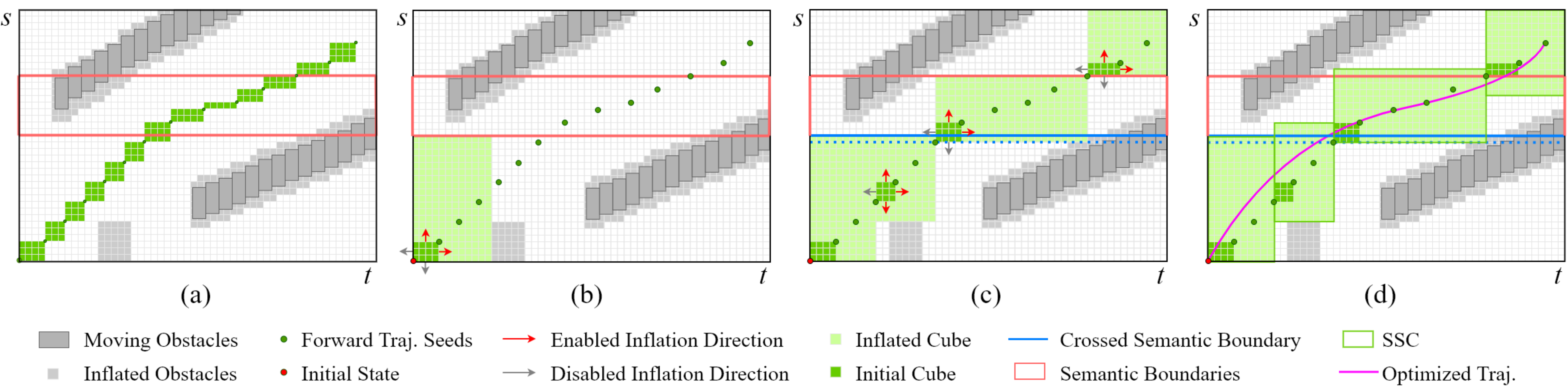}
	\caption{Illustration of a toy example of the SSC generation algorithm in the $st$ domain. There is a speed limit which takes effect between the two \textit{orange} boundaries, as shown in (a). To begin, the first initial cube is inflated until the two inflation directions touch the semantic boundary and the obstacle, as shown in (b). Next, the last seed in the first cube and the first seed outside the first cube are picked out to construct the second initial cube, as shown in (c). The inflation for the second initial cube terminates at the semantic boundary. Then for the third initial cube, the inflation direction opposite to the entry direction is disabled. After the cube inflation, a cube relaxation process is applied depending on the constraints associated and the free-space, as shown in (d).}\label{fig:cube_inflation}
	\vspace{-0.3cm}
\end{figure*}

\section{Spatio-temporal Semantic Corridor}\label{sec:ssc_generation}
\subsection{Semantic Elements And Fren\'{e}t Frame Representation}
We deal with an $slt$ 3-D configuration space which consists of the longitudinal direction $s$, the lateral direction $l$ and the time $t$. The longitudinal and lateral directions are with respect to a Fren\'{e}t frame, which is a dynamical reference frame constructed from the reference lane. Typically, the reference lane is extracted from the route information provided by a route planner, as illustrated in Fig.~\ref{fig:overview}. For an unstructured environment where there is no lane available, the reference lane can also be provided by a path planner~\cite{dolgov2010path}.

Rather than generating the corridor in Cartesian coordinates, we adopt the Fren\'{e}t frame representation since most of the semantic elements are associated with the lane geometry. For example, speed limits, traffic lights and stop signs are typically associated with a certain longitudinal range of a lane. Moreover, since human-like driving behavior can typically be decoupled into lateral movements and longitudinal movements, modeling the free-space in these two directions is a more natural representation than modeling free-space in Cartesian coordinates. Time is another necessary dimension since many semantic elements are time-indexed. For instance, the predicted trajectory is time-profiled and can be regarded as a series of spatio-temporal obstacles.

Two typical examples of projecting the semantic elements to a Fren\'{e}t frame are depicted in Fig.~\ref{fig:cover} and Fig.~\ref{fig:workflow}, respectively. Diverse kinds of semantic elements can be generally divided into two categories: obstacle-like and constraint-like semantic elements. We elaborate on this in the following.

\subsubsection{Obstacle-like semantic elements}
Many semantic elements have the physical meaning that a certain portion of the $slt$ domain is not allowed to be driven in. For example, static obstacles can be viewed as obstacles across whole time axes, and dynamic obstacles can be viewed as a series of static obstacles in the time domain according to the predicted trajectory, while a red light can be rendered as an obstacle occupying a particular longitudinal position and time period. After rendering obstacle-like semantic elements to the $slt$ domain, the configuration space is a 3-D occupancy grid.

\subsubsection{Constraint-like semantic elements}\label{subsubsec:constraint_like_semantic_elements}
Apart from the obstacle-like semantic elements, many semantic elements represent dynamical constraints or time constraints. For example, speed limits and stop signs can be viewed as velocity constraints. There are also semantic elements which pose time constraints. For instance, when crossing lanes, the total time of the lane change should not be unreasonably long.

We propose a unified representation, i.e.,~\textit{semantic boundaries}, for all the constraint-like semantic elements. For instance, a speed limit can be regarded as the velocity constraint applied to a longitudinal range $[s_{\text{begin}}, s_{\text{end}}]$, where $s_{\text{begin}}$ and $s_{\text{end}}$ are the two semantic boundaries. The lane change duration constraint can be regarded as a time constraint applied to the lateral range $[d_{\text{begin}}, d_{\text{end}}]$ of the current lane. Essentially, the semantic boundaries represent where a certain semantic element starts and stops taking effect.

Note that there is a minor difference in terms of the ``hardness'' of the constraints. Specifically, the constraints posed by traffic rules (e.g., speed limit) are hard constraints which should be followed without any compromise. Other constraints (e.g., lane change duration constraint) are required for a natural human-like behavior and there is no universal quantitative description of such constraints. We take the difference into account during the corridor generation process (Sect.~\ref{sec:ssc_generation}).

\subsection{Semantic Corridor Generation}
As outlined in Algo.~\ref{algo:corridor_gen}, the generation process consists of seed generation (Line $3$), cube inflation (Line $4$), constraint association (Line $5$) and cube relaxation (Line $6$).

\subsubsection{Seed Generation}
The seeds of the semantic corridor are generated by projecting the forward simulated states of the behavior planner to the $slt$ configuration space. Since the forward simulated states are discretized, the feasibility of the corridor generation process depends on the complexity of environments and seed resolution. To guarantee the success of the corridor generation process, we require the initial cubes constructed from consecutive seeds to be collision-free (Fig.~\ref{fig:cube_inflation} (a) and Line $6$, Algo.~\ref{algo:corridor_inflation}). In practice, this clearance requirement is reasonable and easy to achieve. For example, for a vehicle travelling at a longitudinal speed of $30$ $m/s$ and a seed resolution of $0.15$ $s$ (similar to \cite{galceran2015mpdmchangept}), the clearance required is roughly $4.5$ $m$, which is much shorter than the emergency braking distance at such a high speed. Therefore, it is reasonable to directly reject the cases which violate the proposed requirement.

The motivation for generating the corridor around the seeds is to fully model topologically equivalent free space, while preserving the same high-level behavior. For example, as shown in Fig.~\ref{fig:cube_inflation} (a), the semantic meaning of the seeds is to pass between the two dynamic obstacles, which is preserved by the corridor generation. Since the motion planner should work with any given initial state, the initial state should also be included in the seeds.
\begin{figure}[t]
	\removelatexerror
	\begin{algorithm}[H]\label{algo:corridor_gen}
		\caption{Semantic Corridor Generation}
		Inputs: forward simulated states $\{x_0,x_1,\ldots,x_t\}$, initial state $x_{\text{des}}$, semantic boundaries $\mathcal{B}$, $slt$ configuration space $\set{E}$\;
		Initializes: seeds $\mathcal{S}^{\text{seed}} = \emptyset$ \;
		$\mathcal{S}^{\text{seed}} \leftarrow \func{SeedGeneration}(\{x_0,x_1,\ldots,x_t\}, x_{\text{des}}) $\;
		$\mathcal{C}^{\text{infl}} \leftarrow \func{CubeInflation}(\mathcal{S}^{\text{seed}}, \mathcal{B}, \set{E})$ \;
		$\mathcal{C}^{\text{infl}} \leftarrow \func{ConstraintAssociation}(\mathcal{C}^{\text{infl}}, \mathcal{B})$ \;
		$\mathcal{C}^{\text{final}} \leftarrow \func{CubeRelaxation}(\mathcal{C}^{\text{infl}}, \set{E})$ \;
	\end{algorithm}
  \vspace{-0.7cm}
\end{figure}
\subsubsection{Cube Inflation with Semantic Boundaries}
The corridor is generated by iterating over the seeds. The seeds which are already contained in the last inflated cube are skipped (Line $4$, Algo.~\ref{algo:corridor_inflation}) since they are topologically equivalent. The initial cubes are generated based on two consecutive seeds, by regarding the two seeds as two cube vertices (Line $5$, Algo.~\ref{algo:corridor_inflation}).

\begin{figure}[t]
	\removelatexerror
	\begin{algorithm}[H]\label{algo:corridor_inflation}
		\caption{$\func{CubeInflation}(\mathcal{S}^{\text{seed}}, \mathcal{B}, \set{E})$}
		Inputs: cube seeds $\mathcal{S}^{\text{seed}}$, semantic boundaries $\mathcal{B}$, $slt$ configuration space $\set{E}$\;
		Initializes: inflated cubes $\mathcal{C}^{\text{infl}} = \emptyset$ \;
		\For{$i=2,\ldots, |\mathcal{S}^{\text{seed}}|$}
		{
			\If{$!\func{IfContainedInLastCube}(s^{\text{seed}}_i, \mathcal{C}^{\text{infl}})$ }
			{
				$ c \leftarrow  \func{GetInitialCubeBySeed}(s^{\text{seed}}_i, s^{\text{seed}}_{i-1})$\;
				\If{$!\func{IfInitialCubeFree}(c, \mathcal{E})$}
				{
					\textbf{return};
				}
				$ \mathcal{D} \leftarrow \func{GetInflDirsBySemBoundaries}(c, \mathcal{B})$\;
				$ c^{\text{infl}} \leftarrow \func{InflateCubeInDirs}(c, \mathcal{D}, \mathcal{B}, \mathcal{E})$\;
				$ \mathcal{C}^{\text{infl}} \leftarrow \mathcal{C}^{\text{infl}} \cup c^{\text{infl}} $
			}
		}
	\end{algorithm}
 \vspace{-0.3cm}
\end{figure}

The key feature of the cube inflation is the consideration of the semantic boundaries (Line $9$, Algo.~\ref{algo:corridor_inflation}). The goal of the cube inflation process is to generate cubes which match the semantic boundaries so that the constraints can be conveniently associated. Specifically, when the initial cube intersects with a certain semantic boundary, the inflation direction opposite to the entry direction is disabled, so that the inflated cube can almost match the semantic boundaries. The inflation alternates among three $slt$ directions for one step of inflation and terminates if this step collides with an obstacle or intersects with a certain semantic boundary. A toy example is provided in Fig.~\ref{fig:cube_inflation} (b) and (c). Since in the optimization (Sect.~\ref{sec:traj_gen}) each cube corresponds to one piece of the trajectory and to preserve convexity we do not explicitly optimize the durations of the pieces, the time upper bound of the current cube should coincide with the time lower bound of the next cube. One may consider optimizing the durations (which is non-convex) and in such case, a further inflation to increase overlapping in the $t$ dimension can be beneficial.

\subsubsection{Cube Relaxation}\label{subsubsec:cube_relaxation}
After the cube inflation process, the inflated cubes almost match the semantic boundaries, as shown in Fig.~\ref{fig:cube_inflation} (c).
However, as mentioned in~\ref{subsubsec:constraint_like_semantic_elements}, some constraints, such as the lane change duration constraint, are soft and extra space should be left for optimization. To this end, we adopt a cube relaxation process to relax the cube boundaries while preserving the hard constraints and collision-free property, as shown in Fig.~\ref{fig:cube_inflation} (d). The maximum margin allowed for the relaxation is systematically determined by the constraints applied to the two consecutive cubes. For example, in the longitudinal direction, the margin can be dermined by velocity matching distance according to the velocity constraints. For the lateral direction (i.e., the lane change case), the margin can be calculated by the allowed fluctuation of lane change duration.

\section{Trajectory Generation With Safety and Feasibility Guarantee}\label{sec:traj_gen}
Given the constraints specified by the SSC, we present an optimization-based trajectory generation method which can find the optimal trajectory within the SSC while satisfying the dynamical constraints.
The optimization problem is also formulated in the Fren\'{e}t frame, which is consistent with the SSC representation.
In~\cite{werling2012mp}, Werling~\textit{et al.} use a quintic monomial polynomial for both the longitudinal and lateral direction based on the optimal control theory. However, the quintic monomial polynomial is not suitable for the optimization in the SSC for the following two reasons: 1) one segment of the polynomial only has limited representation ability and may fail to represent a highly constrained maneuver required by the SSC, and 2) the monomial basis polynomial is not well suited to problems with complex configuration space obstacles and dynamical constraints. In previous works on monomial basis polynomial trajectories~\cite{werling2012mp, fan2018baidu}, the constraints are only enforced/checked on a finite set of sampled points. However, this method may fail to detect collision between sample points, and thus cannot provide any guarantee on safety and feasibility.

In this paper, we remove the above two limitations by using a piecewise B\'{e}zier curve for the two-dimensional trajectory (i.e., the longitudinal direction $s(t)$ and lateral direction $l(t)$) along the reference lane. The reason for using the piecewise B\'{e}zier curve is its convex hull property and hodograph property~\cite{gao2018online}.

\subsection{B\'{e}zier Basis and Its Properties}
A degree-$m$ B\'{e}zier curve $f(t)$ is defined on a fixed interval $t\in[0,1]$ by $m+1$ control points as follows,
\vspace{-0.2cm}
\begin{equation}
	f(t) = p_0b_m^0(t)+p_1b_m^1(t)+\cdots+p_mb_m^m(t) = \sum_{i=0}^{m}p_i \cdot b_m^i(t),
\end{equation}
where $p_{i}$ denotes the control point and $b_m^i(t) = {m \choose i}t^i\cdot(1-t)^{m-i}$ is the Bernstein basis. Denote the set of control points $[p_{0},p_{1},\ldots,p_{m}]$ as $\mathbf{p}$.

The \textit{convex hull} property is suitable for the problem of constraining the curve in a convex free-space. Specifically, the B\'{e}zier curve $f(t)$ is guaranteed to be entirely confined in the convex hull supported by the control points $\mathbf{p}$. In other words, by constraining $\mathbf{p}$ inside the convex free-space, the resulting curve is guaranteed to be collision-free.

\begin{figure}[t]
	\centering
	\includegraphics[width=0.35\textwidth]{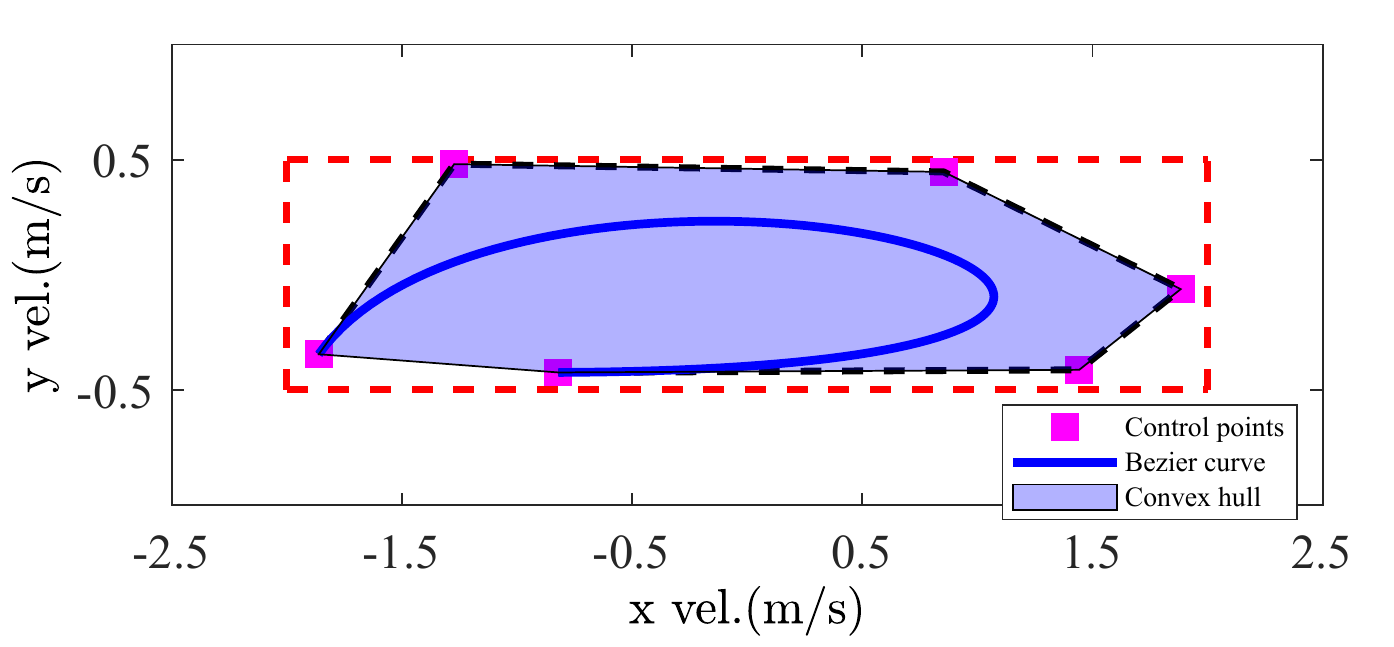}
	\caption{Illustration of using the convex hull property to constrain a velocity profile inside a feasible region (dashed red lines).}\label{fig:convex_hull_example}
	\vspace{-0.5cm}
\end{figure}

The \textit{hodograph} property facilitates constraining high-order derivatives of the B\'{e}zier curve, which is useful for enforcing dynamical constraints. By the hodograph property, the derivative of a B\'{e}zier curve $\frac{df(t)}{dt}$ is another B\'{e}zier curve with control point $p_i^{(1)} = m\cdot (p_{i+1}-p_{i})$. By applying the convex hull property on the derivative B\'{e}zier curve, the entire dynamical profile of the original curve $f(t)$ can be confined within a given dynamical range, as shown in Fig.~\ref{fig:convex_hull_example}.

\subsection{Piecewise B\'{e}zier Curve Representation}
In this paper, we adopt a piecewise B\'{e}zier curve representation with each piece associated with one cube of the SSC. Accordingly, the $j$-th segment of an $n$-segment piecewise B\'{e}zier trajectory in one dimension $\sigma \in \{s, l\}$ is given by
\begin{equation} \label{eq:qp_formulation}
	f_j^{\sigma}(t)=
	\begin{cases}
		\alpha_{1} \cdot \sum_{i=0}^{m}p^{1}_i \cdot b_m^i(\frac{t-t_0}{\alpha_{1}}),     & t\in[t_0, t_1]     \\
		\alpha_{2} \cdot \sum_{i=0}^{m}p^{2}_i \cdot b_m^i(\frac{t-t_1}{\alpha_{2}}),     & t\in[t_1, t_2]     \\
		\quad \quad \quad \vdots \quad \quad \quad \quad \,                               & \quad \vdots       \\
		\alpha_{n} \cdot \sum_{i=0}^{m}p^{n}_i \cdot b_m^i(\frac{t-t_{n-1}}{\alpha_{n}}), & t\in[t_{n-1}, t_n], \\
	\end{cases}
\end{equation}
where $p_i^{j}$ denotes the $i$-th control point of the $j$-th segment and $t_0, t_1, \ldots, t_n$ are the time stamps of the start point and end point for each segment.
Since the B\'{e}zier curve is defined on the fixed interval $[0,1]$ while the trajectory duration for each segment may vary, we introduce a scaling factor $\alpha_{j}$ for each segment according to its duration, similar to~\cite{gao2018online}.

Similar to~\cite{werling2012mp}, we minimize the cost function given by the time integral of the square of the jerk. Specifically, the cost $J_j$ of the $j$-th segment can be written as,
\vspace{-0.2cm}
\begin{equation}
	J_j = w_s \int_{t_{j-1}}^{t_j} \left(\frac{d^3 f^s(t)}{d t^3} \right)^2 dt + w_l \int_{t_{j-1}}^{t_j} \left(\frac{d^3 f^l(t)}{d t^3} \right)^2 dt,
\end{equation}
where $w_s$ and $w_l$ denote the weight for the control cost of the longitudinal direction and lateral direction, respectively. The objective is simple and invariant given different combinations of semantic elements thanks to the SSC, which allows the formulation to easily adapt to different traffic configurations.

Denote by $y^{\sigma}_{j}(t)$ the non-scaled B\'{e}zier curve in the interval $[0,1]$ with $\mathbf{p}_j$ as the control points. Let $u = \frac{t-t_{j-1}}{\alpha_j}$ denote the normalized time of the non-scaled B\'{e}zier curve, the cost of the $j$-th segment on dimension $\sigma$ can be rewritten using the non-scaled $y^{\sigma}_{j}(t)$ as follows,
\vspace{-0.2cm}
\begin{equation}
	J^{\sigma}_j  =\!\int_{0}^{1}\!\!\alpha_j \cdot \left(\frac{d^3(\alpha_j \cdot y_j^{\sigma}(t))} {d (u \cdot \alpha_j)^3} \right)^2\!d u \nonumber = \frac{1}{\alpha_j^3} \cdot \mathbf{p}_j^{\text{T}} \mathbf{Q} \mathbf{p}_j,
\end{equation}
where $\mathbf{Q}$ is the Hessian matrix of the non-scaled B\'{e}zier curve. We omit the detailed calculation of $\mathbf{Q}$ for brevity.

\subsection{Enforcing Safety and Dynamical Constraints}
In this paper, we adopt a quintic ($m$=5) piecewise B\'{e}zier curve as the trajectory parameterization. According to the hodograph property, the $k$-th derivative of the non-scaled B\'{e}zier curve $\frac{d^k y^{\sigma}_j(t)}{dt^k}$ is supported by control points $\mathbf{q}^{\sigma, (k)}_j$ which can be calculated by induction as follows,
\vspace{-0.2cm}
\begin{equation}
q^{\sigma, (0)}_{j,i} = p^j_{i}, q^{\sigma, (k)}_{j,i} = \frac{m!}{(m-k)!}(q^{\sigma, (k-1)}_{j,i+1} - q^{\sigma, (k-1)}_{j,i}).
\end{equation}
Based on this property, the $k$-th-order derivatives at the boundaries of $f^{\sigma}_j(t)$ can be expressed as
\vspace{-0.2cm}
\begin{equation}\label{eq:start_end_derivative}
	\frac{d^k f^{\sigma}_j(t_{j-1})}{dt^k} = \alpha_j^{1-k} \cdot q^{\sigma, (k)}_{j,0},\frac{d^k f^{\sigma}_j(t_j)}{dt^k} = \alpha_j^{1-k} \cdot q^{\sigma, (k)}_{j,m},
\end{equation}
respectively. Moreover, by further applying the convex hull property, we can constrain the entire derivative profile of $f^{\sigma}_j(t)$ using the following sufficient condition,
\vspace{-0.1cm}
\begin{equation}\label{eq:sufficient_bound}
	\beta_{j,-}^{\sigma,(k)} \! \leq \! \alpha_j^{1-k}\cdot q^{\sigma, (k)}_{j,i} \! \leq \! \beta_{j,+}^{\sigma,(k)} \!, \forall i \Rightarrow \!	\beta_{j,-}^{\sigma,(k)} \leq \frac{d^k f^{\sigma}_j(t)}{dt^k} \! \leq \! \beta_{j,+}^{\sigma,(k)},
\end{equation}
where $\beta_{j,-}^{\sigma,(k)}$ and $\beta_{j,+}^{\sigma,(k)}$ denote the lower and upper bound on dimension $\sigma$ for the $k$-th derivative of the $j$-th segment.

\subsubsection{Desired state constraints}
First of all, the generated trajectory should start from the given initial state $[\sigma^{(0)}_{t_0},\sigma^{(1)}_{t_0},\sigma^{(2)}_{t_0}]$ and terminate at the given goal state $[\sigma^{(0)}_{t_n},\sigma^{(1)}_{t_n},\sigma^{(2)}_{t_n}]$ for $\sigma\in \{s,l\}$, where $\sigma^{(k)}_{t}$ denotes the $k$-th-order derivative at time $t$. Specifically, this requires enforcing equality constraints for the first and last segment as follows,
\vspace{-0.1cm}
\begin{equation}
\frac{d^k f^{\sigma}_0(t_0)}{dt^k} = \sigma^{(k)}_{t_0},  \quad \quad  \frac{d^k f^{\sigma}_n(t_n)}{dt^k} = \sigma^{(k)}_{t_n},
\end{equation}
where $k=0,1,2$. By applying Eq.~\ref{eq:start_end_derivative}, these constraints can be written as linear equality constraints w.r.t. $\mathbf{p}$.

\subsubsection{Continuity constraints}
The generated trajectory should be continuous for all the derivatives up to the $k$-th order at all the connecting points between two consecutive pieces. The continuity constraints between the $j$-th segment and the $j+1$-th segment can be written as
\vspace{-0.1cm}
\begin{equation}
	\frac{d^k f^{\sigma}_j(t_j)}{dt^k} = \frac{d^k f^{\sigma}_{j+1}(t_j)}{dt^k},
\end{equation}
where $k=0,1,2,3$. By applying Eq.~\ref{eq:start_end_derivative}, these constraints can also be written as linear equality constraints w.r.t. $\mathbf{p}$.

\subsubsection{Free-space constraints}
To guarantee the generated trajectory is collision-free, we constrain each segment of the trajectory within the corresponding cube. The free-space constraint of the $j$-th segment on dimension $\sigma$ can be enforced by using the sufficient condition (Eq.~\ref{eq:sufficient_bound}) under $k=0$, where $\beta_{j,-}^{\sigma,(0)}$ and $\beta_{j,+}^{\sigma,(0)}$ represent the position bounds on dimension $\sigma$ given by the shape of the cube.

\subsubsection{Dynamical constraints}
To comply with the environment semantics and dynamical feasibility constraint, we enforce the constraints on the derivatives of trajectories by using the sufficient condition (Eq.~\ref{eq:sufficient_bound}), where $k=1,2$. The physical meaning is that the maximum lateral/longitudinal velocity and acceleration is constrained. Summarizing all the linear equality and inequality constraints, the overall formulation can be written as a QP, which can be solved efficiently using off-the-shelf solvers (such as OOQP). Although Eq.~\ref{eq:sufficient_bound} is a sufficient condition, in practice we find it does not result in over-conservative behavior, as shown in Sect.~\ref{sec:experimental_results}. In the case that no feasible solution can be found, the error is fed back to the behavior layer for further reaction.

\begin{figure*}[t]
	\centering
	\begin{subfigure}[b]{0.24\textwidth}
		\centering
		\includegraphics[width =0.95\textwidth]{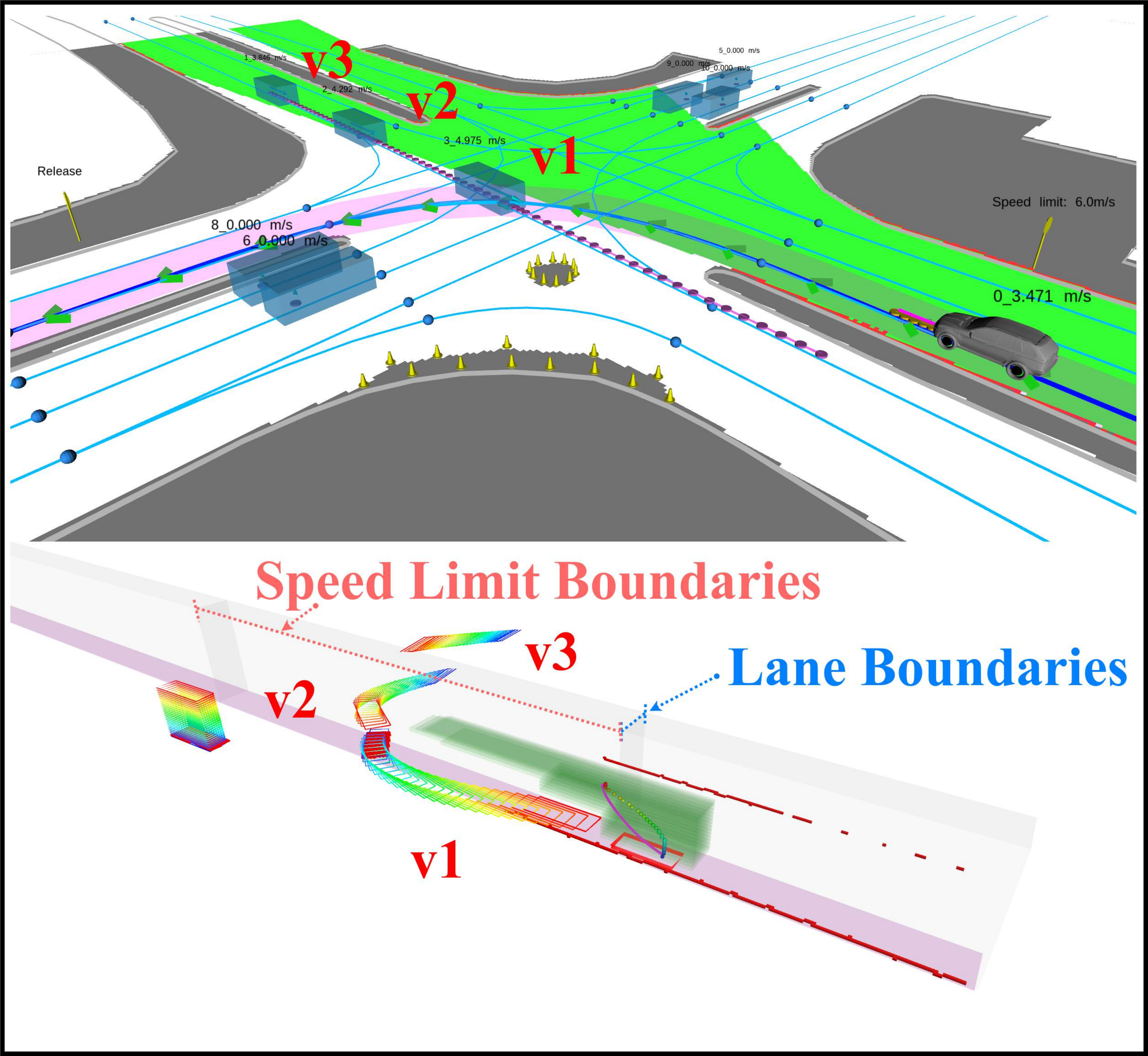}
		\caption{Approaching the intersection}
	\end{subfigure}
	\begin{subfigure}[b]{0.24\textwidth}
		\centering
		\includegraphics[width =0.95\textwidth]{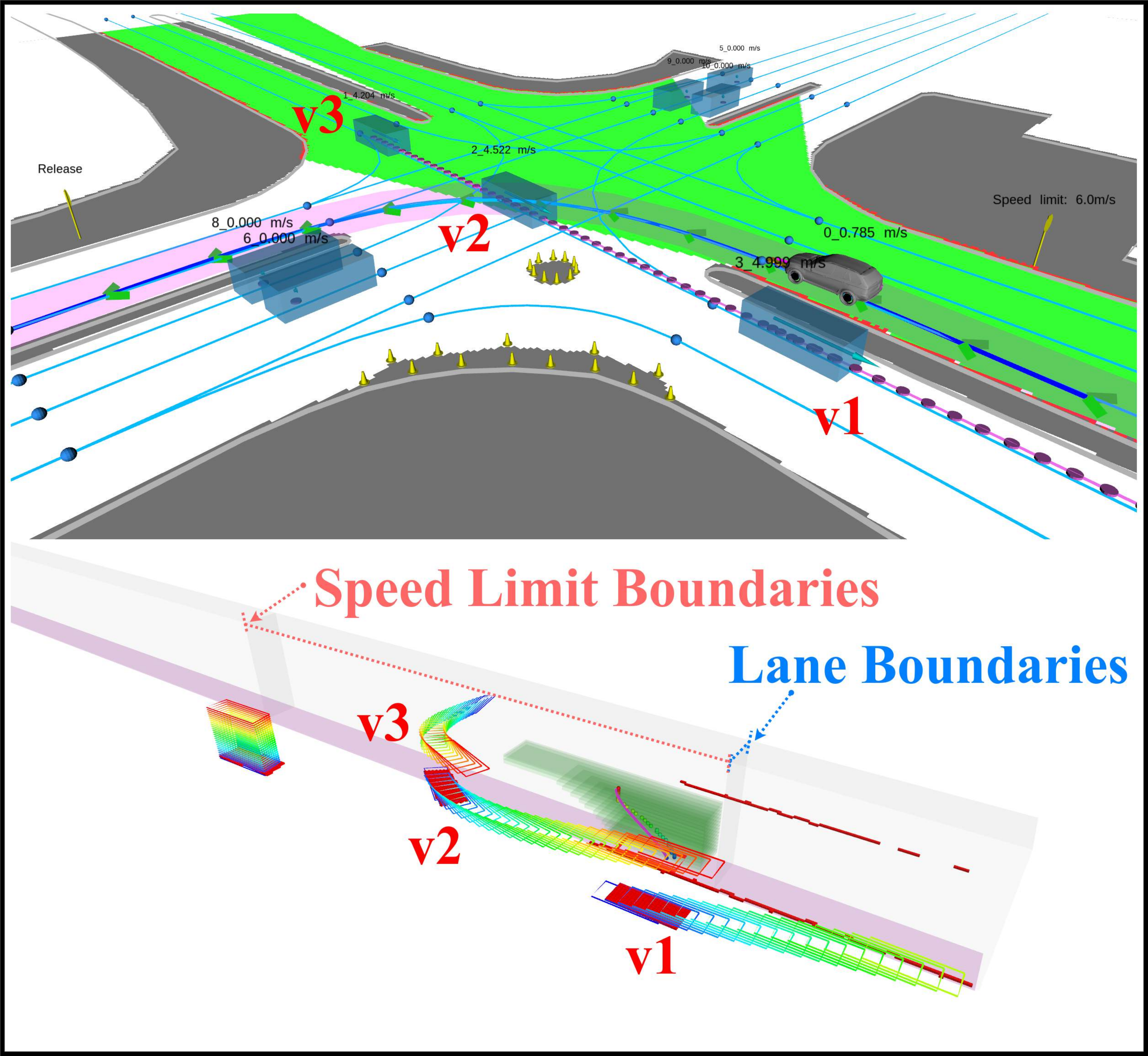}
		\caption{Reducing speed to wait}
	\end{subfigure}
	\begin{subfigure}[b]{0.24\textwidth}
		\centering
		\includegraphics[width =0.95\textwidth]{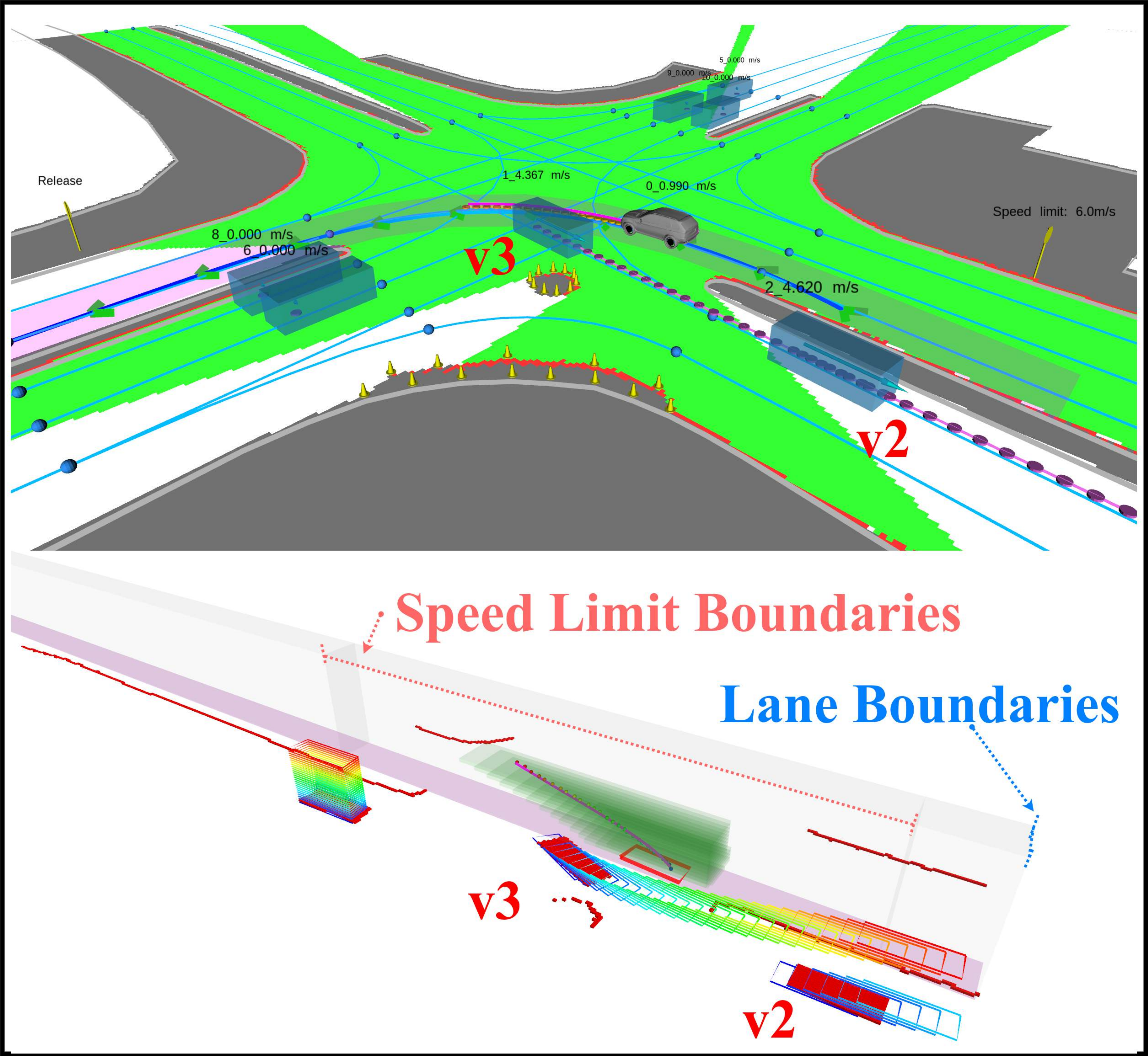}
		\caption{Accelerating to pass}
	\end{subfigure}
	\begin{subfigure}[b]{0.24\textwidth}
		\centering
		\includegraphics[width =0.95\textwidth]{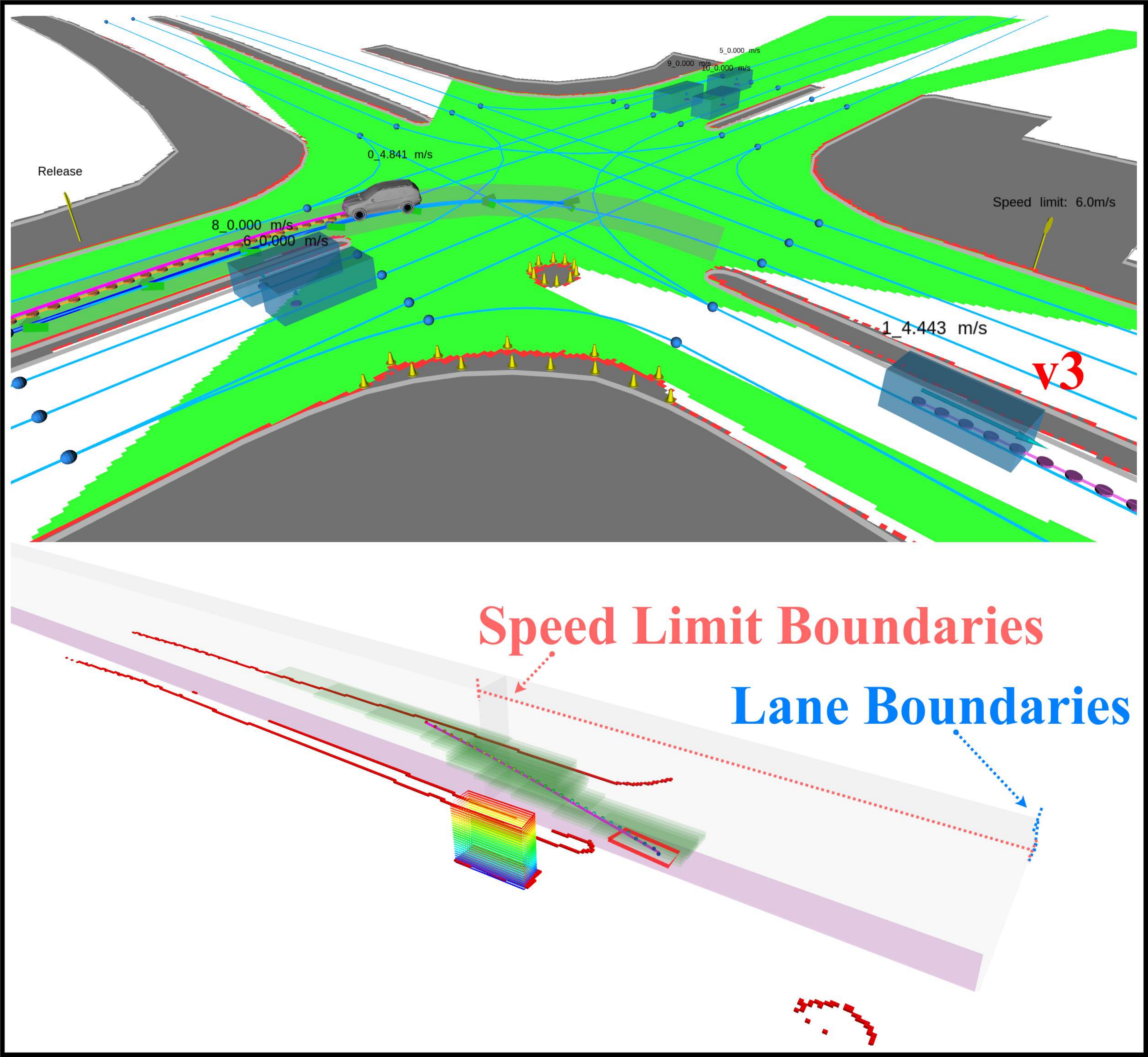}
		\caption{Completing the left turn}
	\end{subfigure}
	\caption{Illustration of an unprotected left turn in a busy urban intersection. When the ego vehicle is approaching the intersection, it finds the left turn is not feasible and it reduces speed to wait. Once feasible, the vehicle quickly accelerates to complete the left turn.}\label{fig:exp_left_turn}
	\vspace{-0.4cm}
\end{figure*}

\section{Experimental Results}\label{sec:experimental_results}
\subsection{Implementation Details}\label{sec:impl_details}
The experiments are conducted in a multi-agent simulation platform, as illustrated in Fig.~\ref{fig:overview}. In the simulation, dynamic agents potentially interact with each other, but the interaction model is unknown to the planner. The ego (our) vehicle only has a limited sensing range for the environment semantics. The route planner finds a random route for the ego vehicle at the beginning of the mission, and the route information of other agents is also unknown to the planner. The prediction method is similar to that in~\cite{ding2018onlinepred} which decouples the problem into behavior prediction and trajectory prediction. A long prediction horizon facilitates accounting for long-term future rewards, which potentially results in a more consistent output compared to using a short prediction horizon. However, the uncertainty also scales with the prediction horizon. Therefore, it is beneficial to characterize the long-term prediction uncertainty, and we provide an attempt in~\cite{ding2019int}. All the test environments are annotated from real satellite maps via QGIS. The planning method proposed in this paper\footnote{\scriptsize{Source code is released at} \url{https://github.com/HKUST-Aerial-Robotics/spatiotemporal_semantic_corridor}.} is implemented in C++11. All the experiments are conducted on a desktop computer equipped with an Intel I7-8700K CPU, and our proposed method can run stably at $20$ Hz.

\subsection{Qualitative Results}
To verify that our proposed method can automatically adapt to different traffic configurations with different semantic elements, we choose three representative test cases.

\subsubsection{Merging into congested traffic due to road construction}
As illustrated in Fig.~\ref{fig:workflow}, this case is used to verify the capability of dealing with road construction, lane change (lane geometry), dynamic obstacles and the speed limit at the same time. The constructed SSC generally encodes the necessary information for optimization. The optimal trajectories are generated without explicitly caring about what types of semantic elements are present.

\subsubsection{Overtaking on an urban expressway}
\begin{figure}[t]
	\centering
	\includegraphics[width=0.48\textwidth]{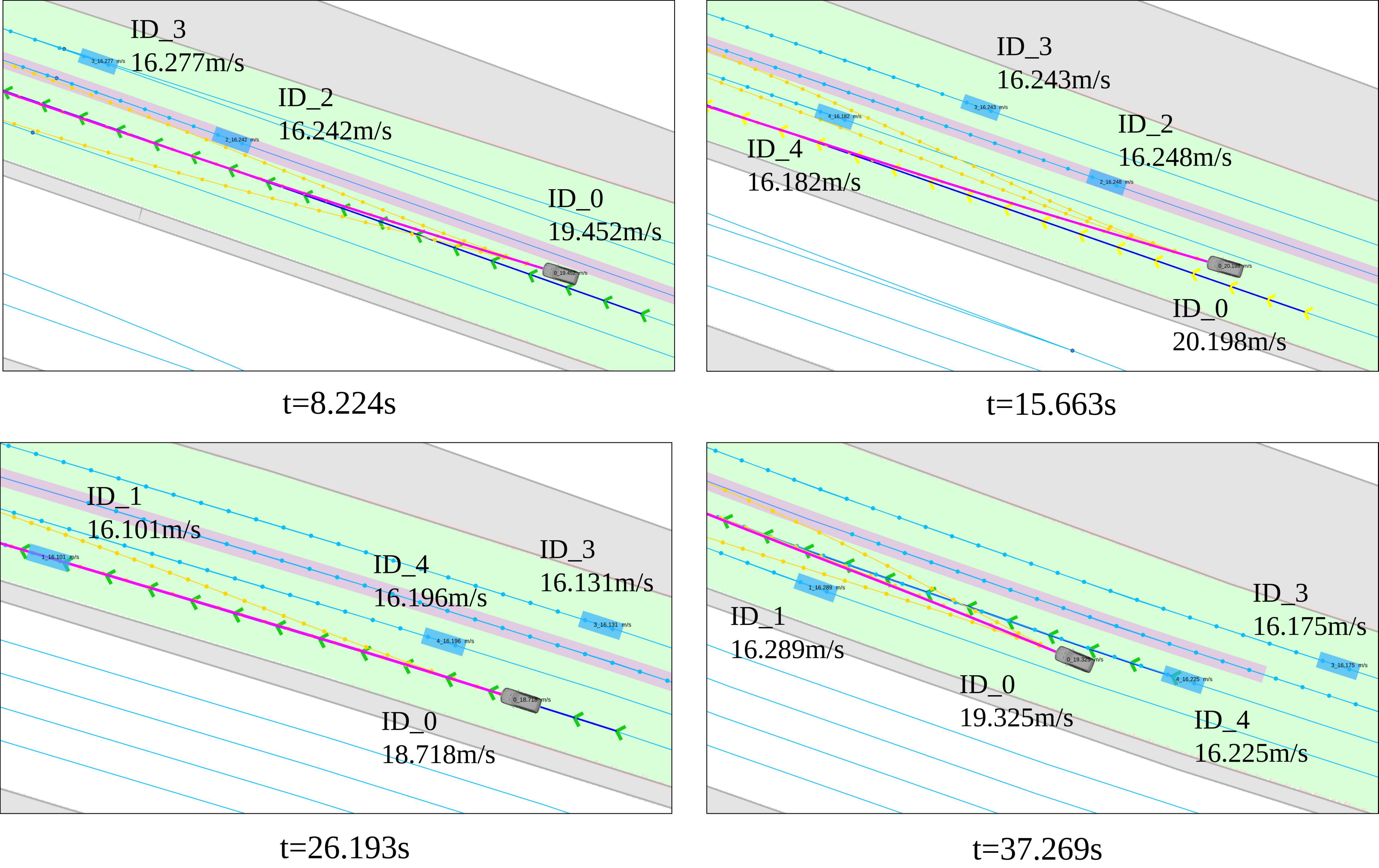}
	\caption{Illustration of overtaking on an urban expressway.}\label{fig:exp_overtake}
	\vspace{-0.5cm}
\end{figure}

This case is to validate the capability of dealing with high-speed traffic. The SSC is shown to be suitable for this time-critical scenario. As illustrated in Fig.~\ref{fig:exp_overtake}, our method conducts a safe and smooth overtaking on an urban expressway with a speed of around $20$ $m/s$. The limitation is that the prediction uncertainty is not sufficiently considered in the current SSC generation process, which is left as important future work.

\subsubsection{An unprotected left turn at an intersection}
This case is used to verify the capability of quickly responding to complex interactions with other agents during traffic negotiation. There is also a speed limit which poses hard speed constraints for the whole interaction process. As shown in Fig.~\ref{fig:exp_left_turn}, our method efficiently finds safe and feasible trajectories so that the vehicle precisely follows the behavior plan and navigates smoothly.

\subsection{Comparisons and Analysis}
We conduct a quantitative comparison with the seminal work~\cite{werling2012mp}, which is based on optimal primitives in the Fren\'{e}t frame. In~\cite{werling2012mp}, the primitives are regularly sampled around a local target state with a certain resolution in the $slt$ domain, and for different behaviors, the strategy for choosing the local target is different.

To conduct a fair comparison, we set up a benchmark track which is annotated from a real satellite map, as shown in Fig.~\ref{fig:benchmark_satellite}. To test the planner's response to semantic elements, we add a red light checkpoint and a speed limit to the track. Moreover, dense obstacles are placed on the track, as shown in Fig.~\ref{fig:benchmark_rviz}, to test the collision avoidance performance.
Since the ego vehicle only has a limited sensing range (around 100 $m$), the collision avoidance task requires frequent replanning. The maximum acceleration and the maximum deceleration are set to $2$ $m/s^2$ and $3$ $m/s^2$, respectively. We use the same behavior planner (MPDM) for both our method and~\cite{werling2012mp} to generate the lane change command. The user-desired velocity is set to $15$ $m/s$ for the behavior planner.

\subsubsection{Collision-avoidance in cluttered environments}\label{subsubsec:collision_avoidance_cluttered}
The first segment of the track is around 320 $m$ from the starting point to the red light. As shown in Fig.~\ref{fig:benchmark_dynamic}, our method can fully utilize the maneuverability of the vehicle and arrive at the red light at $42$ $s$, about $14$ seconds earlier than~\cite{werling2012mp}.
Moreover, our acceleration profile is smoother while staying within the dynamical limit. The reason is that our SSC representation models the continuous solution space, while the baseline method suffers from discretization and limited state space coverage. We observe that the benefit of using the corridor representation is more obvious in the cluttered environments since many primitives of the baseline method become infeasible in this case.

\subsubsection{Precise stop with a high entry speed}
There is a red light checkpoint in the middle of the track, as shown in Fig.~\ref{fig:benchmark_satellite}, and the vehicle needs to complete a precise stop with a high entry speed. As shown in Fig.~\ref{fig:benchmark_dynamic}, our method can reach the precise stop with an entry speed of $13$ $m/s$ while the max deceleration is strictly bounded inside the dynamic range. However, for the baseline method~\cite{werling2012mp}, a stopping mode is needed to fix the local target state so that the replanning process can consistently reach the target boundary condition. If we do not manually fix the local target state and dynamically calculate it based on zero desired velocity, the initial sampled stopping trajectory may not be sampled in later replanning due to a minor change of the target state. This may cause rolling, as shown in Fig.~\ref{fig:benchmark_dynamic}. In contrast to~\cite{werling2012mp}, our method explicitly enforces the stopping boundary condition and achieves a precise stop.

\subsubsection{Collision avoidance under a low speed limit}
\begin{figure}[t]
  \centering
  \begin{subfigure}[b]{0.19\textwidth}
  	\centering
    \includegraphics[width =0.95\textwidth]{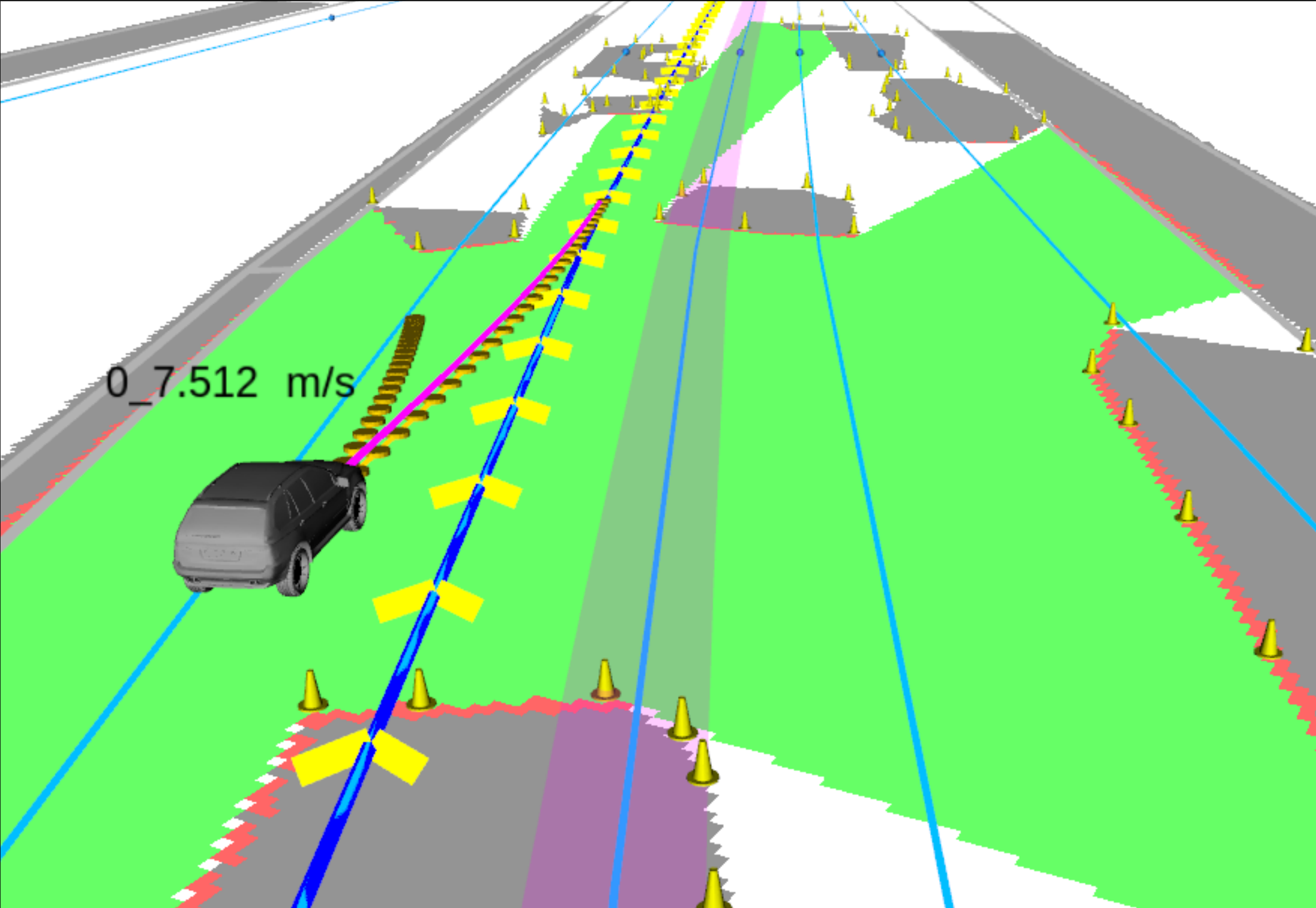}
    \caption{Benchmark track}\label{fig:benchmark_rviz}
  \end{subfigure}%
  \begin{subfigure}[b]{0.19\textwidth}
  	\centering
    \includegraphics[width =0.95\textwidth]{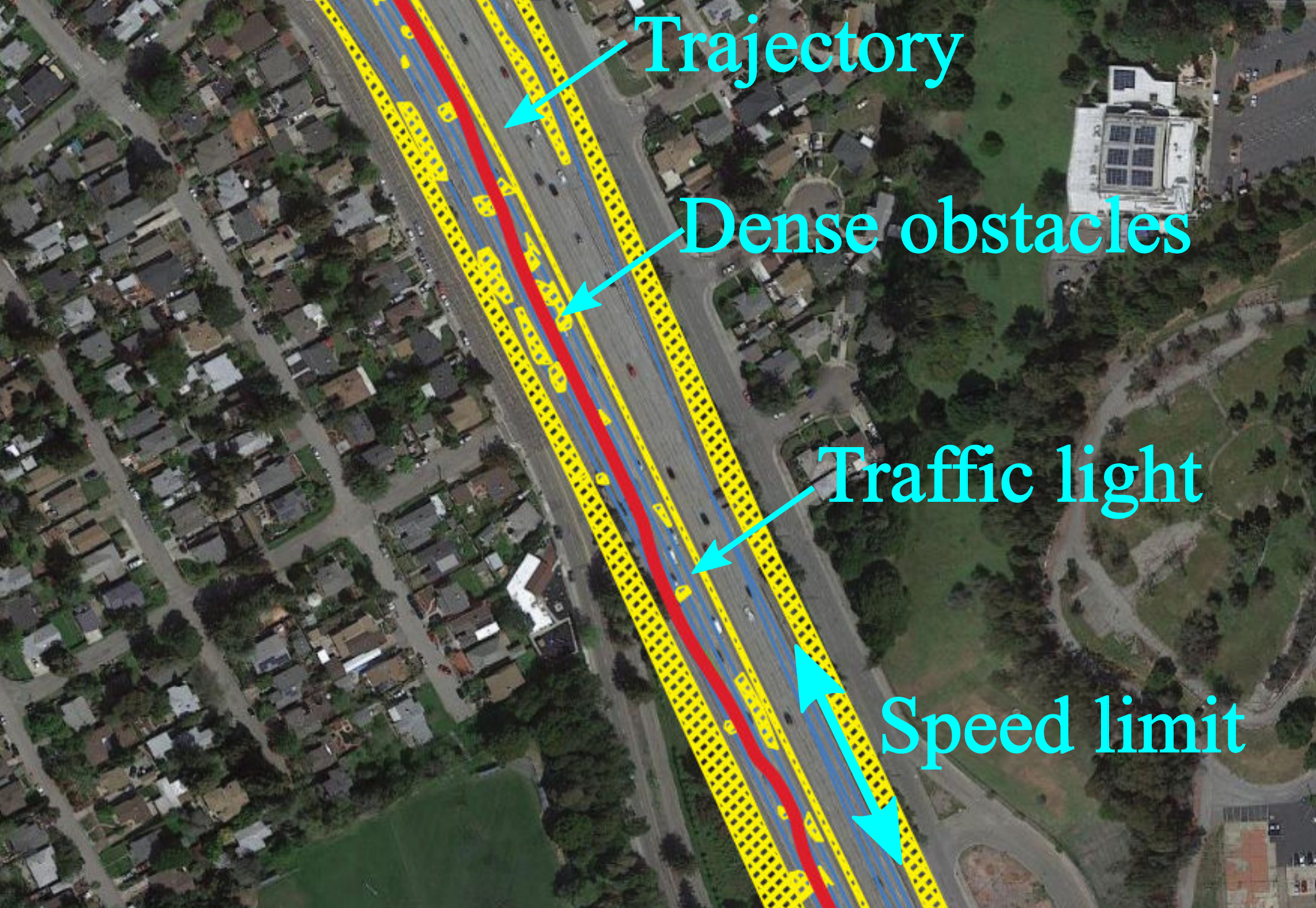}
    \caption{Satellite map}\label{fig:benchmark_satellite}
  \end{subfigure}
  \begin{subfigure}[b]{0.41\textwidth}
  	\centering
		\includegraphics[width =\textwidth]{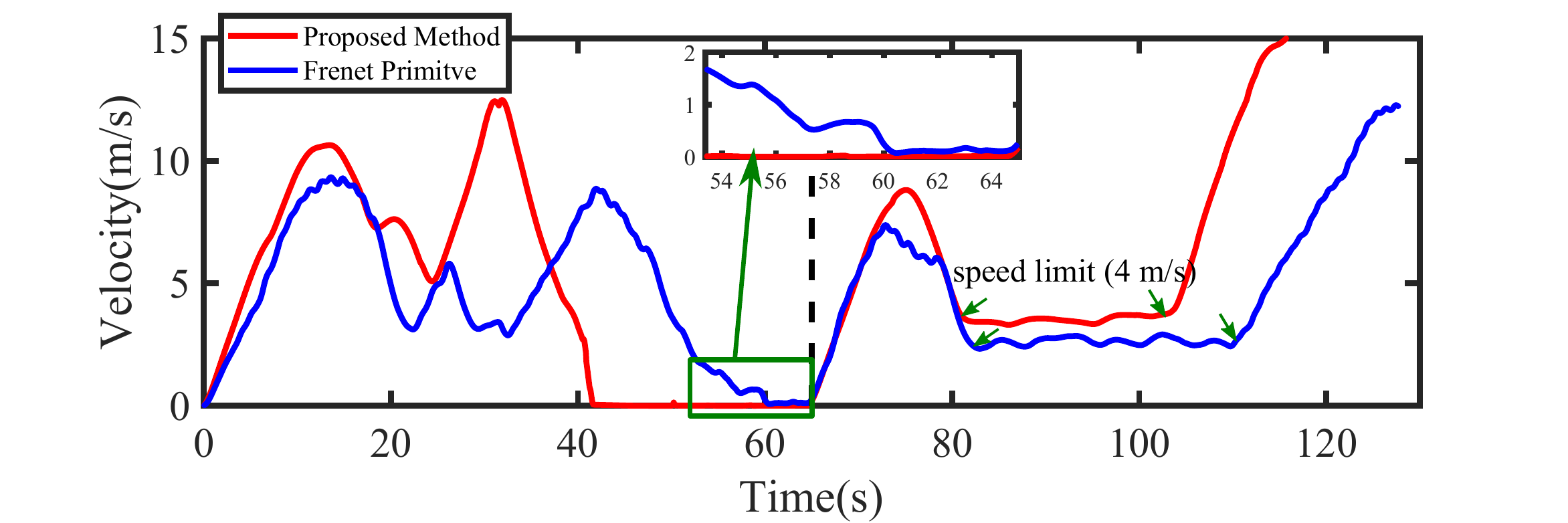}
    \includegraphics[width =\textwidth]{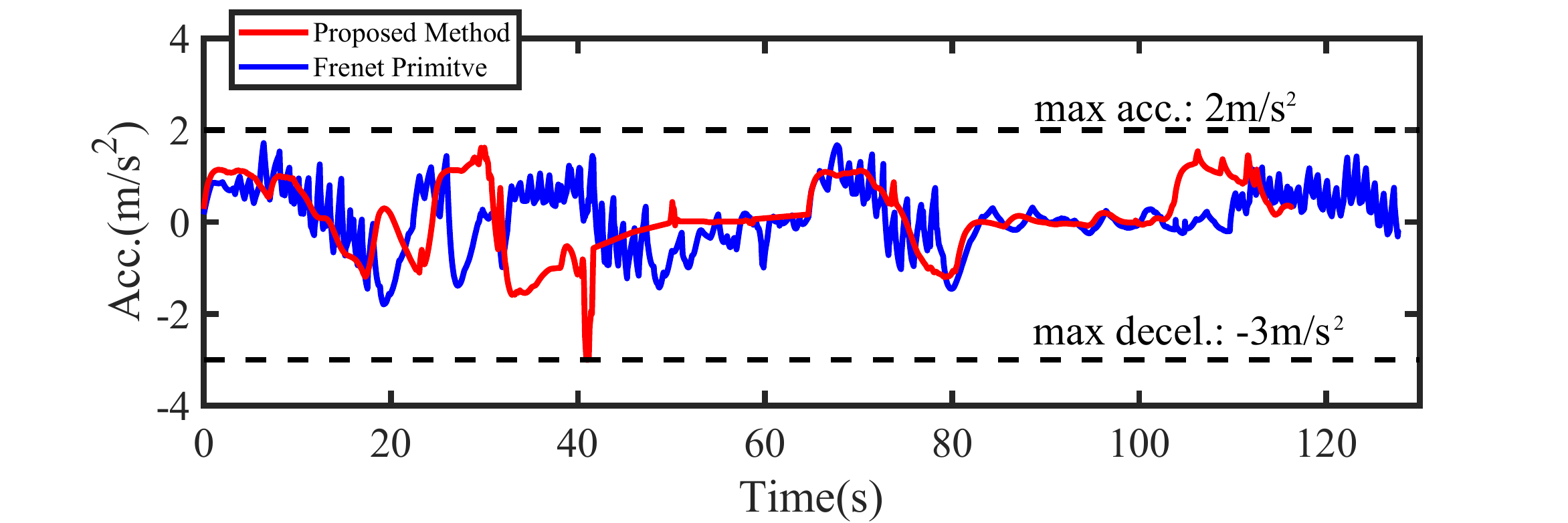}
    \caption{Comparison of the dynamic profile}\label{fig:benchmark_dynamic}
	\end{subfigure}
	\caption{Illustration of the comparison on a benchmark track.}\label{fig:benchmark}
	\vspace{-0.5cm}
\end{figure}

In addition to the study of high-speed collision-avoidance, we are also interested in the low-speed performance. To test this, a $4$ $m/s$ speed limit is placed on the track, as shown in Fig.~\ref{fig:benchmark_satellite}. As depicted in Fig.~\ref{fig:benchmark_dynamic}, our method strictly follows the speed limit.

We also conduct experiments in which obstacles are placed in an online manner (see our video for details).

\section{Conclusion and Future Work}\label{sec:conclusion}
In this paper, we propose a trajectory generation framework for complex urban environments. Our main contribution is twofold. First, we present an SSC structure which copes with an arbitrary combination of semantic elements in a unified way. Second, we present a trajectory optimization formulation which guarantees the safety and feasibility of the output trajectory. The proposed method is extensively analyzed using various traffic configurations and complex semantic elements. The main limitation is that the prediction uncertainty and interaction uncertainty are not sufficiently modeled, which is the research direction we are currently working on~\cite{ding2019int}. Moreover, we find that the B\'{e}zier curve is also useful for non-linear trajectory optimization for AVs.

\end{document}